\documentclass[sigconf,screen]{acmart} 
\usepackage{lscape}

\AtBeginDocument{%
  \providecommand\BibTeX{{%
    \normalfont B\kern-0.5em{\scshape i\kern-0.25em b}\kern-0.8em\TeX}}}

\setcopyright{acmcopyright}
\copyrightyear{2024}
\acmYear{2024}
\acmDOI{XXXXXXX.XXXXXXX}
\usepackage{dsfont}
\usepackage{multirow}

\acmConference[AI-ML Systems '23]{}{25-28 Oct, 2024}{Bangalore, India}

\DeclareMathOperator{\argmax}{argmax} 
\let\ALP \mathcal

\newcommand{\Na}{\mathbb{N}}

\newcommand{\ind}[1]{\mathds{1}\{#1\}}
\begin{document}

\title{Maximizing Success Rate of Payment Routing using Non-stationary Bandits}

\author{Aayush Chaudhary}
\affiliation{%
  \institution{Dream11}
  \city{Mumbai}
  \country{India}
}
\email{c.aayush@alumni.iitg.ac.in}

\author{Abhinav Rai}
\affiliation{%
  \institution{Dream11}
  \city{Mumbai}
  \country{India}}
\email{abhinav@dream11.com}

\author{Abhishek Gupta}
\affiliation{%
  \institution{Ensemble Control Inc. \&}
  \institution{The Ohio State University}
  \city{Columbus, OH}
  \country{USA}
}
\email{gupta.706@osu.edu}

\renewcommand{\shortauthors}{Chaudhary, et al.}

\begin{abstract}
This paper discusses the system architecture design and deployment of non-stationary multi-armed bandit approaches to determine a near-optimal payment routing policy based on the recent history of transactions. We propose a Routing Service architecture using a novel Ray-based implementation for optimally scaling bandit-based payment routing to over 10000 transactions per second, adhering to the system design requirements and ecosystem constraints with Payment Card Industry Data Security Standard (PCI DSS). We first evaluate the effectiveness of multiple bandit-based payment routing algorithms on a custom simulator to benchmark multiple non-stationary bandit approaches and identify the best hyperparameters. We then conducted live experiments on the payment transaction system on a fantasy sports platform Dream11. In the live experiments, we demonstrated that our non-stationary bandit-based algorithm consistently improves the success rate of transactions by 0.92\% compared to the traditional rule-based methods over one month. \end{abstract}






\maketitle
\section{Introduction}
Electronic payment systems are an essential component for enabling transactions on modern platforms. Payment processors are crucial in ensuring that transactions are processed efficiently and accurately. However, routing payments to the most appropriate payment processors can be challenging, mainly because the success rates are highly non-stationary and vary across the payment processors \cite{zanzot2010systems,ciobotea2017,zimmerman2019routing}. The traditional rule-based approaches may not be sufficient to optimize the complex business metrics of the problem. In recent years, machine learning and reinforcement learning approaches have shown promising results in various domains, including routing payments \cite{adyen2020,linkedin2023,payu2018}. The fantasy sports platform of Dream11 has specific user behaviour that is notable: 
\begin{enumerate}
    \item Strict upper bound on transactions per second (TPS) through a payment gateway. 

    \item Specific time windows experience very high transaction volume, up to 100x times the usual traffic. 

    \item The platform has long-term obligations to meet the minimum number of transactions with each payment gateway. 
\end{enumerate}
As a result of the contractual obligations, the platform needs to design a learning algorithm that balances the need for meeting the short-term upper bound on the TPS and maintaining the long-term lower bound on the TPS. In addition, the learning algorithm has to work with minimal computational overhead. 


In this research paper, we address the challenge of optimizing payment routing in non-stationary environments, a problem that has not been effectively addressed by existing methods relying on rule-based algorithms or traditional machine-learning techniques. \newline \textit{Summary of contributions:} In this work, the following contributions are made to the field of routing payments in the fintech ecosystem:
\begin{itemize}
    \item We assess bandit-based \textit{sequential decision-making algorithms} in dynamic environments, demonstrating their superiority over baseline methods in online A/B tests. This leads to a 0.92\% increase in overall payment success rates in one month and an impressive 10\% improvement during significant payment gateway performance declines, highlighting the algorithm's adaptability and effectiveness. It' also establishes a strong correlation between simulator-based performance assessments and real-world online metrics.

    \item A robust simulator that helps us conduct comprehensive \textit{simulations using real-world payment data} from Dream11 to evaluate the performance of our proposed framework. This empirical analysis provides insights into the practical effectiveness of our approach; our simulations demonstrate the superiority of our approach in handling non-stationary environments.
    
    \item We propose a \textit{scalable Ray-based implementation} of the routing service. The underlying service architecture’s balance of low latency, high throughput, scalability, and cost-effectiveness makes it well-suited for handling large concurrent transaction volumes.

\end{itemize}

The proposed approach builds upon the previous work in the field of non-stationary bandit algorithm \cite{garivier2011upper,ciobotea2017,wei2021nonstationary,bygari2021ai}. In the traditional bandit algorithm, the environment is assumed to be stationary \cite{auer2002finite}. To deal with a non-stationary environment, typically, the historical information beyond a certain point in the past is either discarded or discounted \cite{garivier2011upper,auer2019achieving,wei2021nonstationary}. We conducted several simulations to determine which algorithm works best for our application and meets the contractual obligations with the payment gateways. We found that the sliding window UCB algorithm with hyperparameter optimized through trial and error performs the best.

Given the non-stationarity of the environment, one has to balance the exploration and exploitation tradeoff continually. As a result, several payment routing algorithms have been proposed in the literature to optimize the efficiency and accuracy of payment routing. Broadly speaking, there are three major approaches to route payments:
\begin{itemize}
    \item In the machine learning approach, features are constructed from the attributes of the transactions. Then, a machine learning algorithm is trained to determine the probability of a successful transaction using each payment gateway. The routing algorithm uses these probabilities to route the payments. The algorithms by \cite{juspay2021,linkedin2023,bygari2021ai} fall within this category.
    \item In the bandit approach, some variant of the non-stationary bandit algorithm is used to route the payments. Several approaches have been described in great detail \cite{ciobotea2017,agrawal2012analysis}. This paper also falls within this class of algorithms; however, the algorithm deployed differs from the algorithms described.
    \item In the contextual bandit approach, each transaction has a context vector associated with it. A contextual bandit approach is then taken to route the payments. The algorithm in \cite{adyen2020} takes this approach.  
\end{itemize}

\section{Problem Formulation and Algorithms}

\paragraph{State space} We assume that at each time $t$, a certain number of customer arrives at the platform and executes a transaction. Let $p_t$ denote the payment processors chosen by the customer. Let $x_t:=(x_{t,g_1}, \ldots, x_{t,g_l})\in[0,1]^{\ALP G}$ denote the success probability of the payment gateways in $\ALP G$. This is the state of the MDP and is not influenced by the action of the platform. We note here that the state of the MDP is unknown and needs to be inferred from the history of transactions.

\paragraph{Action Space} Depending on the payment processor selected by the customer $p_t$, the transaction needs to be routed through the gateways in the set $\ALP G_{p_t}$.

\paragraph{Reward} When the platform selects the payment gateway $g_t\in \ALP G_{p_t} $, the payment is routed through $g_t$. The platform receives the reward signal $r_t\in\{0,1\}$ within 15-30 seconds. Here, $r_t = 1$ implies the payment was successful, and $r_t = 0$ otherwise.   

\paragraph{Historical Information} The sequence of transactions carried out by the platform before time $t$ generates a history 
\begin{align*}
    \ALP H_t = \{(p_k,g_k,r_k)\}_{k=1}^{t-1}.
\end{align*}
This history is used to estimate the success rate of each processor-gateway pair. 

\paragraph{Estimate of the Success Rate}  For each gateway $g$ and time $t$, let $\tau_{t,g}$ denote the stopping time such that there were at least $W$ attempted transactions through the gateway $g$ in the time interval $\{\tau_{t,g},\tau_{t,g}+1,\ldots,t-1\}$. This is defined as
\begin{align*}
    \tau_{t,g} = \max\Bigg\{\tau\in\Na: & \sum_{k=\tau}^{t-1} \ind{g_k = g}\geq W\Bigg\}
    \label{eq:tau}
\end{align*}
We let $N_{t,g}$ and $\hat N_{t,g}$ be defined as
\begin{align*}
    N_{t,g} = \sum_{k = \tau_{t,g}}^{t-1} \ind{g_k = g},\qquad  \hat N_{t,g} = \sum_{k = \tau_{t,g}}^{t-1} \alpha^{t-k}\ind{g_k = g}.
\end{align*}
Now, the success rate for the gateway $g$ is estimated as
\begin{align*}
    y_{t,g} = \frac{\sum_{k = \tau_{t,g}}^{t-1} r_k\ind{g_k = g}}{N_{t,g}}
\end{align*}
Now, the discounted success rate for the gateway $g$ is estimated as
\begin{align*}
    \hat y_{t,g} =\frac{\sum_{k = 0}^{t-1} \alpha^{t-k}r_k\ind{g_k = g}}{\hat N_{t,g}}
\end{align*}
\paragraph{Policy} As in the non-stationary bandit policies, we must continually explore and exploit. For each time $t$, we compute and store $y_{t,g}$ and $\hat y_{t,g}$ in the memory. We define five policies:
\begin{align*}
  \text{$\epsilon$ Greedy }  & \pi_t(\ALP H_t) = \begin{cases}
       \argmax_{g\in\ALP G_{p_t}}y_{t,g} & \text{ w.p. }1-|\ALP G_{p_t}|\epsilon \\
       g\in\ALP G_{p_t} & \text{ w.p. }\epsilon
    \end{cases},\\
    \text{SW-UCB }  & \pi_t(\ALP H_t) = \argmax_{g\in\ALP G_{p_t}} y_{t,g} + c_1 \sqrt{\frac{1}{N_{t,g}}},\\
    \text{SW-BG }  & \pi_t(\ALP H_t) = \argmax_{g\in\ALP G_{p_t}} y_{t,g} + c_1 \sqrt{\frac{1}{N_{t,g}}}Z_{t,g},\\
    \text{D-UCB }  & \pi_t(\ALP H_t) = \argmax_{g\in\ALP G_{p_t}}\hat y_{t,g} + c_1 \sqrt{\frac{1}{\hat N_{t,g}}},\\
    \text{D-BG }  & \pi_t(\ALP H_t) = \argmax_{g\in\ALP G_{p_t}}\hat y_{t,g} + c_1 \sqrt{\frac{1}{\hat N_{t,g}}}Z_{t,g},
\end{align*}
where $Z_{t,g}$ is Gumbel(0,1) random variable and independent of the past realizations and $c_1$ is a constant hyperparameter.

Additionally, we also implemented a discounted version of the Thompson sampling algorithm. This algorithm discounts the past posterior distribution and appropriately updates the posterior distribution using the new reward signal described below. The prior and posterior distribution is assumed to be the Beta distribution, whose parameters are updated as follows:
\begin{enumerate}
    \item Sample $\theta_{t,g} \sim \text{Beta}(\lambda_{t,g}, \gamma_{t,g})$ for each gateway $g$.
    \item The discounted Thompson sampling policy is
    \begin{equation*}
        g_t := \pi_t(\ALP H_t) = \argmax_{g\in\ALP G_{p_t}} \theta_{t,g}
    \end{equation*}
    and observe the reward $r_t$.
    \item Update the parameters of the Beta distribution representing the success rate:
    \begin{align*}
        \lambda_{t+1,i}, \gamma_{t + 1,i} =
        \begin{cases}
            \lambda_{t,i}\alpha+r_t, \gamma_{t,i} \alpha + (1-r_t), & \text{if } i = g_t, \\
            \lambda_{t,i}, \gamma_{t,i}, & \text{if } i \neq g_t,
        \end{cases}
    \end{align*}
    where $\alpha$ is the discount factor.
\end{enumerate}

In the policies described above, note that $W$ and $\alpha$ are the hyperparameters for sliding window algorithms and discounted algorithms respectively. Thus, we devised families of parameterized bandit policies, which we implemented in the routing context. We ran a grid search to tune the hyperparameters for each learning algorithm in the simulations for the live production system. 

To conduct live experiments, we needed to identify the correct software system architecture to enable testing across the cohort of users. We explain the system architecture below. 

\begin{figure*}[!t]
  \centering
  \includegraphics[width=0.8\linewidth]{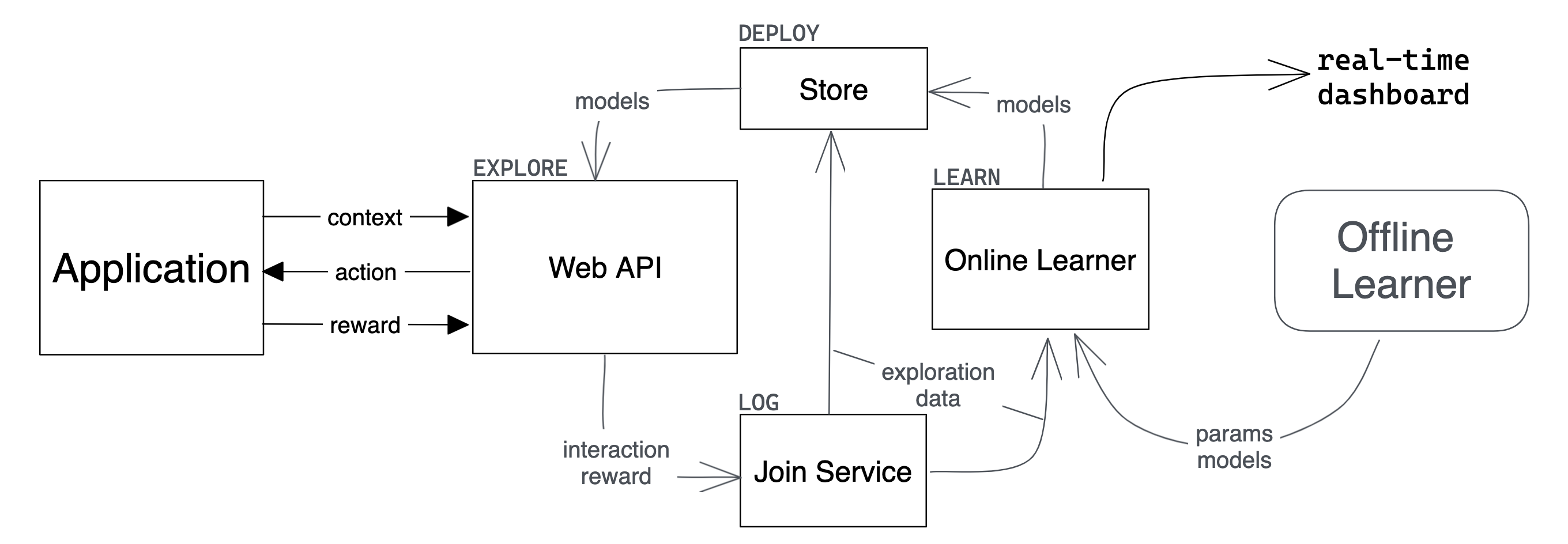}
  \caption{Service architecture of the decision service taken from \cite{agarwal2017making}.}
  \label{fig:sa}
\end{figure*}

\section{System Architecture}

\subsection{System Requirements}

The proposed system design focuses on critical requirements for successful implementation and functionality.

\begin{enumerate}
    \item Scalability: The system architecture must efficiently handle traffic surges ranging from \textbf{5 to 20 times} the usual load within \textbf{10 to 15 minutes} during peak times. Key considerations include achieving low latency with response times below \textbf{5 milliseconds} for real-time interactions and high throughput to support over \textbf{10,000 transactions per second (TPS)}.
    
    \item Reliability: To minimise downtime and service disruptions, the system should guarantee high availability, with an uptime of \textbf{99.99\%}.
    
    \item Security: Routing service is security-sensitive, and it is subject to the Payment Card Industry Data Security Standard (PCI DSS) \cite{pcidoc},\cite{pciweb}. Failing these requirements, the company is subject to strict penalties \cite{pcinoncomplaince}. 
    
    \item Maintainability: Automated deployment and monitoring mechanisms will be implemented to ensure continuous system availability and timely issue resolution.
    
    \item Usability: The system specifically cater to two stakeholders \textbf{Data Scientists}, who can run experiments, and \textbf{Business Users}, responsible for monitoring and updating configurations. 
    \begin{enumerate}
     \item {Experimentation:} The system should also allow for concurrent experiments to facilitate continuous learning and model performance evaluation.
     \item {Rate-limiting:} The system should be able to prevent overloading payment gateways, respecting their transaction rate constraints. 
    \end{enumerate}
    \item \textbf{Cost:} The architecture aims to minimize the utilization of compute (cores), memory, and database queries, optimizing cost-effectiveness without compromising system performance.
\end{enumerate}

The proposed system is expected to deliver a highly scalable, reliable, secure, maintainable, user-friendly, and cost-effective payment processing solution by adhering to these critical design requirements.

\subsection{Bandit System Requirements}
We design our production system to incur low technical debt using the decision service architecture from \cite{agarwal2017making}.  
\begin{enumerate}
\item Explore: Inference of the bandit algorithm to assign scores to payment gateways based on business agreements with merchants and banks.
\item Log: This component generates the reward data (success/failure).  
\item Learn: This module performs online policy training based on the rewards accrued and returns the updated policy.  
\item Deploy: The updated policy is deployed based on the experiment configuration.
\end{enumerate}

The delay in receiving the acknowledgement of payment success is a common issue in payment routing. Despite this delay, the success probability of the payment does not typically change significantly within 15 seconds. To address this issue, the simulation assumes instant knowledge of the reward signal and acknowledges that this assumption may need to be overcome, as discussed in Section [\ref{sec:conc}].

\subsection{Ecosystem Constraints}
The existing ecosystem imposes the following constraints:
\begin{enumerate}
\item Microservices: The system has to be designed as micro-services that can be deployed independently and interact with the payments service
\item Routing Requests: The main payments micro-service will send an HTTP request with the partial context. 
\item Rewards: Payment request outcomes (success/failure) are available on a Kafka topic/HTTP request. Querying the database is not an option due to the PCI DSS Compliance requirements\cite{pcidoc}.
\item Deployment zone: The routing service has to be deployed on the same soiled zone as the payments service.
\end{enumerate}

\subsection{System Design for Bandit Implementation}
Three system design choices were considered:

\begin{enumerate}
    \item \textbf{Classical Approach:}
    \begin{itemize}
        \item Web API: Using a Python-based HTTP service and a Database. The Database stores all the reward logs. For each request, the HTTP service hits the database to fetch stored models and serves the application with the optimal routing suggestion (Figure 1)
        \item Scheduled Jobs: for the online learning process, a scheduler is used which reads the rewards logged in the above web APIs database and calculates the expected rewards and updates the same to the database
        \item Database: A key-value store like Redis  or an RDBMS like MySQL can be used based upon the scalability and performance requirements \cite{seghier2021performance}.
    \end{itemize}

    \item \textbf{Clipper Based Approach} \cite{crankshaw2017clipper}:

\begin{itemize}
    \item Frontend Service: Entry point for applications to send routing requests, handles batching, and directs queries to model containers for high throughput.
    
    \item Model Selection Service: Decides which model/version should handle a given query based on historical and real-time performance data.
    
    \item Model Containers: Docker containers encapsulating individual ML models, providing a standardized interface for frontend service interactions.
    
    \item Management Service: Offers APIs for model management tasks and handles model container scaling based on load.
    
    \item Monitoring and Logging Service: Collects performance metrics, logs, and feedback for insights into system and model performance, facilitating debugging and optimization.
    
    \item Cache Service: Caches frequent query results to reduce the load on model containers and improve response times.
\end{itemize}

    \item Actor-Based System with Shared Memory (Ray-based Implementation \cite{moritz2018ray}):
    \begin{itemize} 
        \item We propose a Ray-based HTTP service, where we use stateful actors for services like model selection, exploration, logging and experimentation. This aproach is deployed as a single unit of Ray cluster (Figure 2). We use plasma store, an in-memory shared memory datastore as the primary storage.     
        \item To make the system more robust when restarting we periodically persist the datasets in the shared memory of the HTTP service to inexpensive storage services like AWS S3 \cite{s3} or Azure Blob \cite{azureblobstorage}. This eliminates the need for a database. 
    \end{itemize}

\end{enumerate}

\begin{figure}[htbp]
  \centering
  \includegraphics[width=\linewidth]{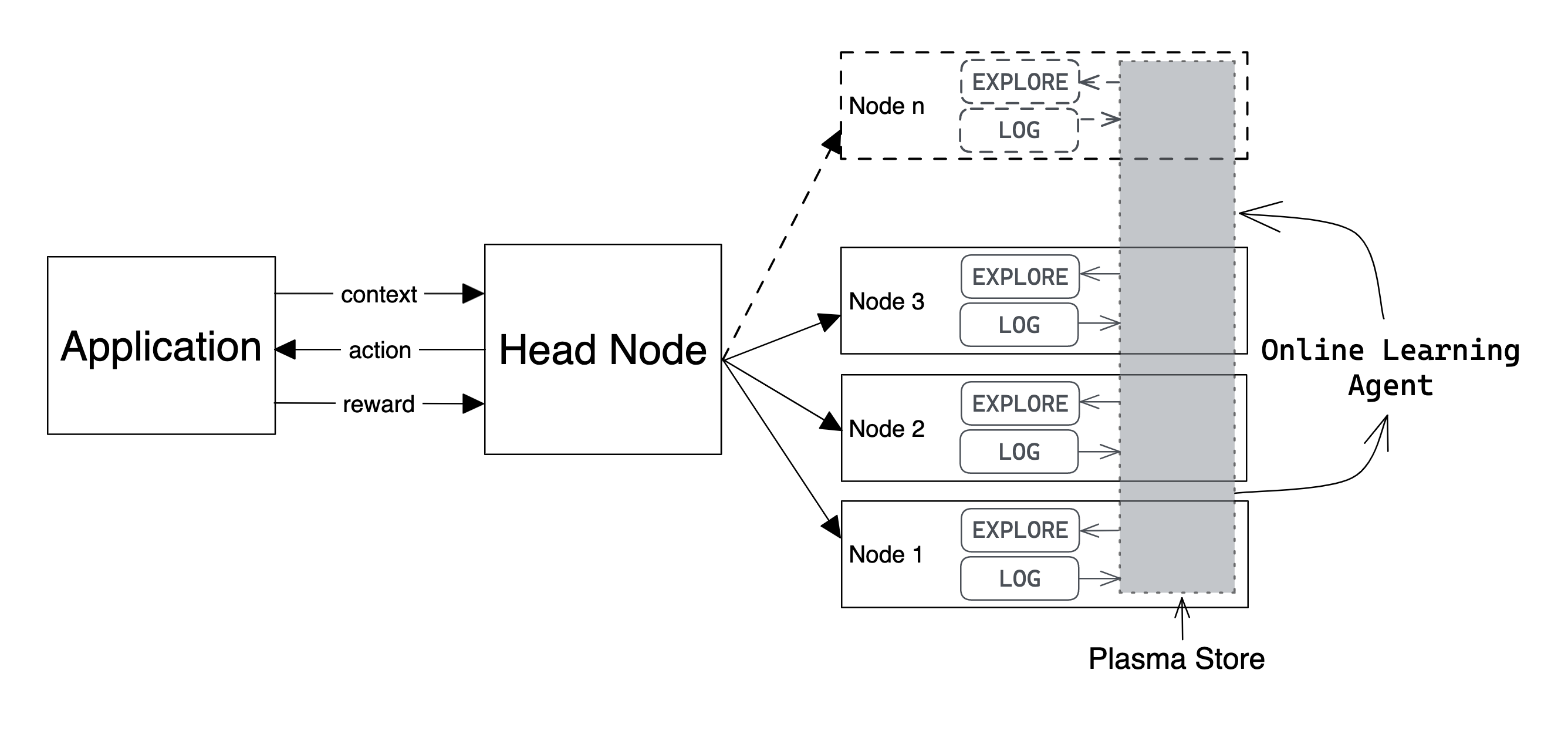}
  \caption{Proposed Routing Service Architecture}
  \label{fig:sa}
\end{figure}

\begin{figure*}[ht]
  \includegraphics[width=0.44\linewidth]{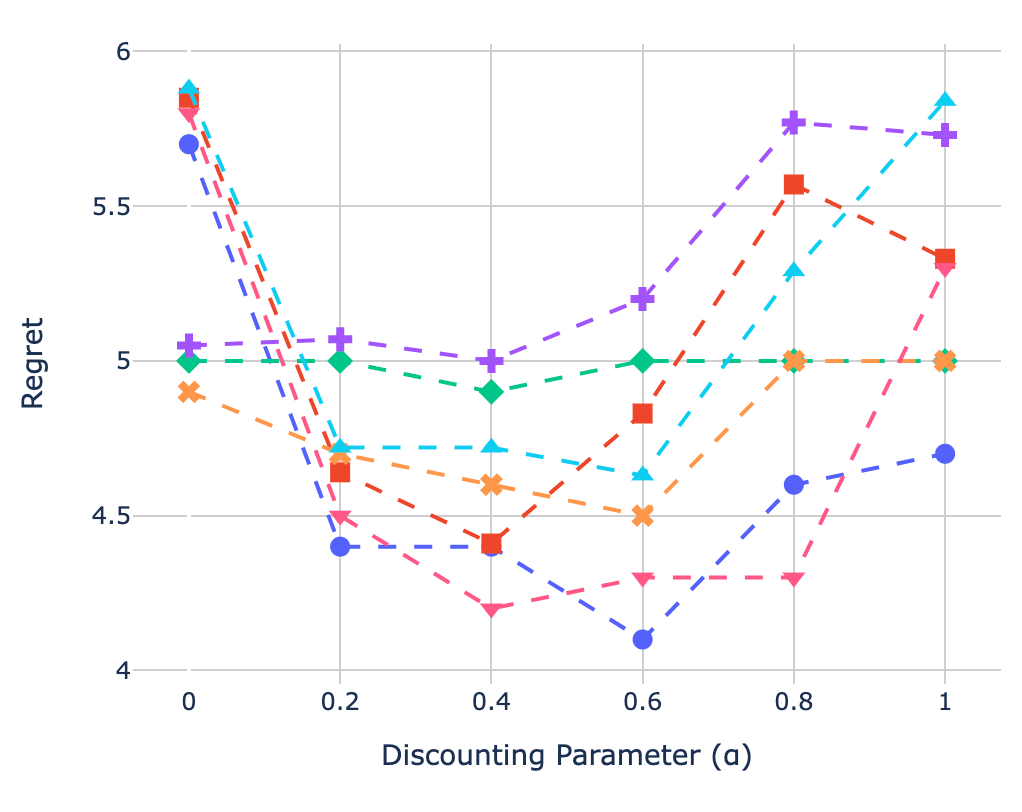}
     \includegraphics[width=0.54\linewidth]{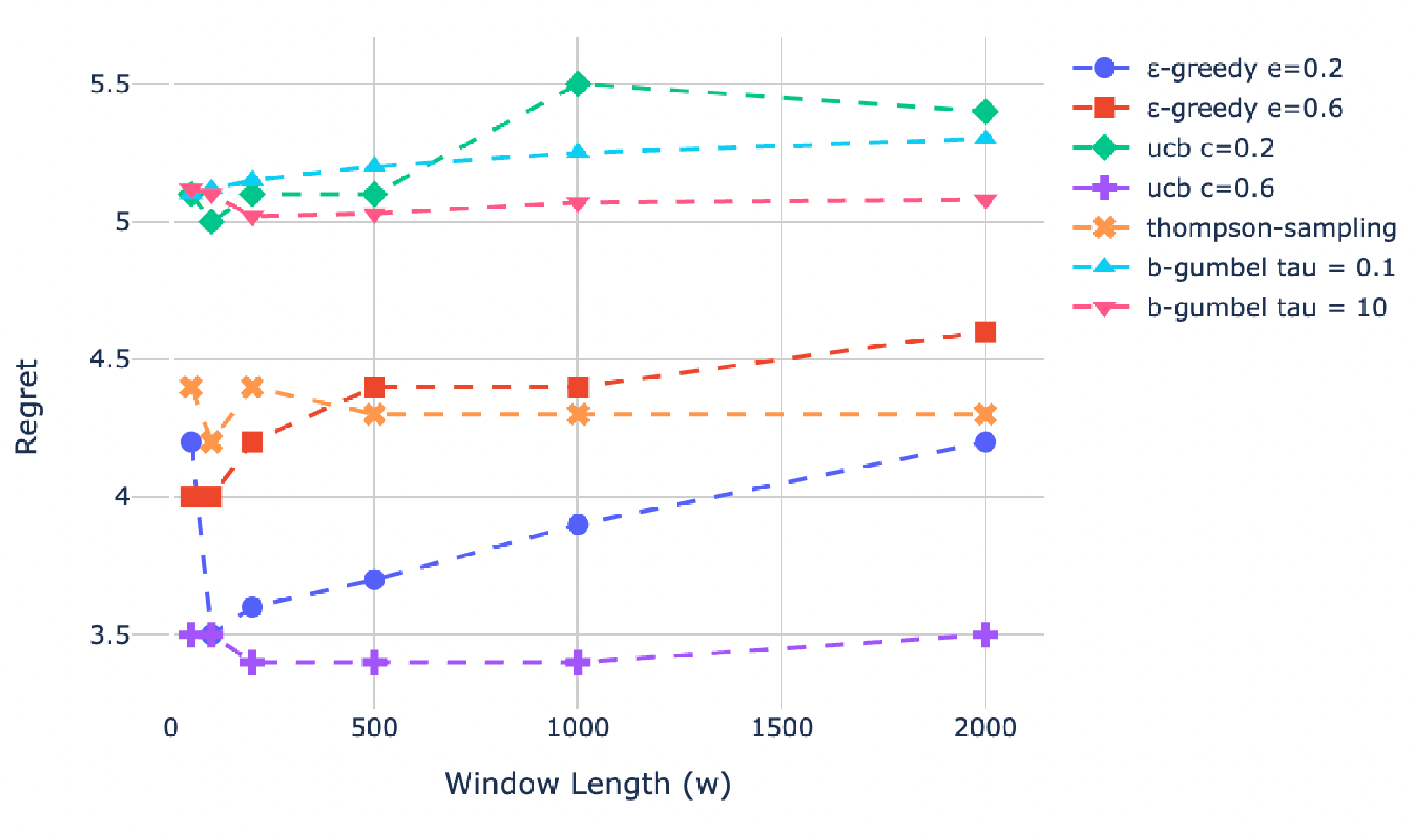}
  \caption{Regret obtained by various non-stationary bandit algorithms tuning dataset. The figure on the left are discounted versions of the algorithms and the figure on the right are sliding window versions.}
  \label{fig:eg_discounting}
\end{figure*}

\begin{figure*}[ht]
  \includegraphics[width=0.95\linewidth]{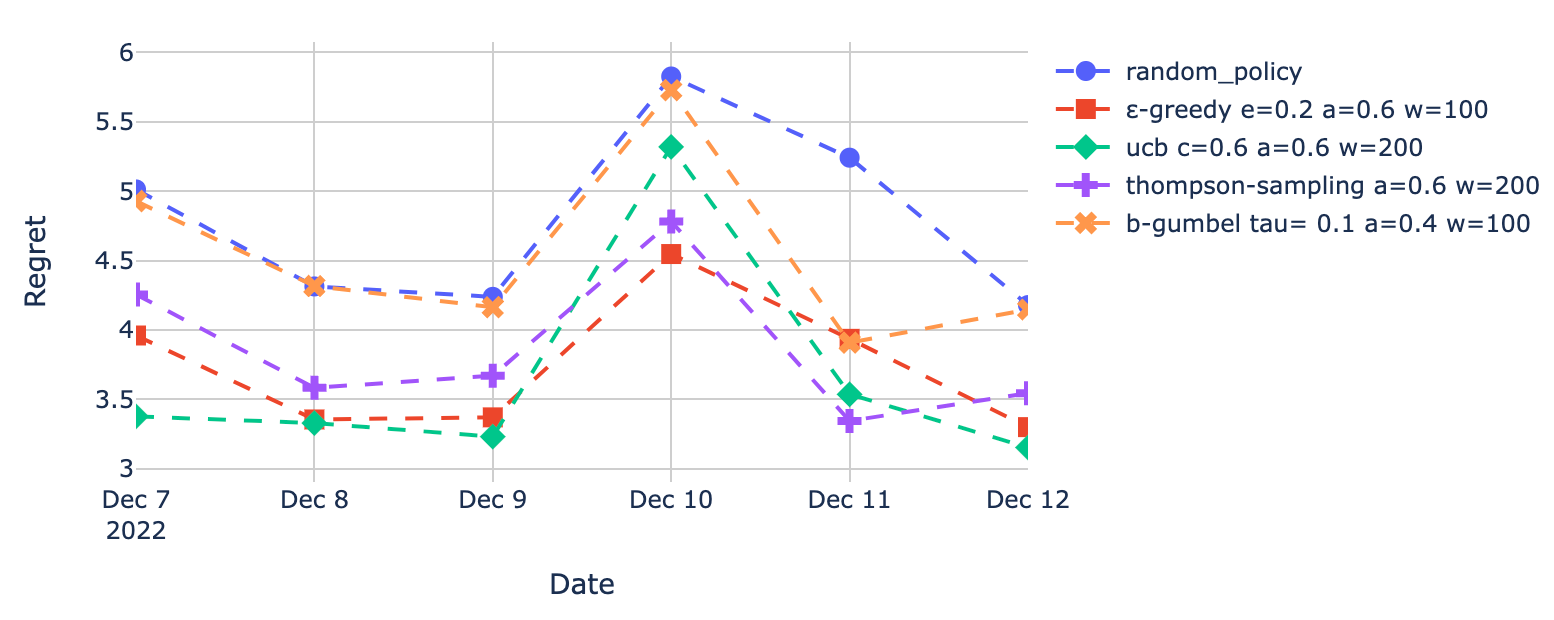}
  \caption{Offline Testing Results: Regret of Bandit Algorithms with Best Hyperparameters on evaluation dataset.}
  \label{fig:offline}
\end{figure*}

\subsection{Tradeoffs in System Design}
There are multiple tradeoffs inherent in the system architectures described above:
\begin{itemize}
 \item The classical approach, although easy to implement, has high latency and is expensive to scale for high throughput.

 \item Clipper-based approach provides low latency bandit inference but adds multiple services, deployment, and PCI \cite{pcidoc} compliance overhead. The rewards update process is also delayed as it uses schedules for online learning. 

 \item Agent-Based System dramatically reduces the deployment, PCI Compliance and MLOps effort as it is deployed as a single service. Using Actors on top of shared memory, this approach offers low latency inference, high throughput, model updates without lag, and efficient handling of scaling and concurrency without data duplication.

\end{itemize}


We used the third method -- the agent-based system -- for our current system implementation. It eliminated the need for a separate database, thereby saving costs significantly. It further reduced the system's latencies by doing away with the overhead linked to the external database connections. The use of an asynchronous agent within this system has allowed us to forgo a job scheduler, as this agent can continuously update the rewards asynchronously within the same service.

The most compelling advantage offered by this method is its ability to scale to a remarkably high level of concurrent requests, a crucial requirement for our system. Hence, despite the inherent tradeoffs, the advantages offered by the third method have far outweighed its limitations, leading us to choose it for our current system implementation.



\section{Experimentation}
\subsection{Dataset}
The routing dataset used in this research comes from over 100M payment transactions conducted on the Dream11 fantasy sports gaming platform for a week in December 2022. We used 1st day’s transactions (“tuning data”) for model validation to decide the optimal parameter for each competing bandit algorithm. Then we ran these algorithms with tuned parameters on a one-week transactions data ( “evaluation data”). The dataset comprises the following attributes for each transaction:

\begin{itemize}
\item \textbf{Id}: A unique identifier for each transaction.
\item \textbf{Amount}: The transaction amount denominated in Indian Rupees (INR).
\item \textbf{Source}: The payment instrument from which the transaction was initiated.
\item \textbf{Terminal}: The identification number of the terminal responsible for processing the transaction.
\item \textbf{Success}: A binary indicator indicating whether the transaction was successful or failed.
\end{itemize}

Two important limitations of the dataset used in our simulation is that we used a week's data in December 2022 to tune the bandit algorithms (i.e., identify the best hyperparameter) and limited representation of diverse payment transactions. Consequently, these factors may influence and restrict the effectiveness and generalizability of routing algorithms trained on this dataset.

\begin{figure*}[ht]
\includegraphics[width=0.9\linewidth]{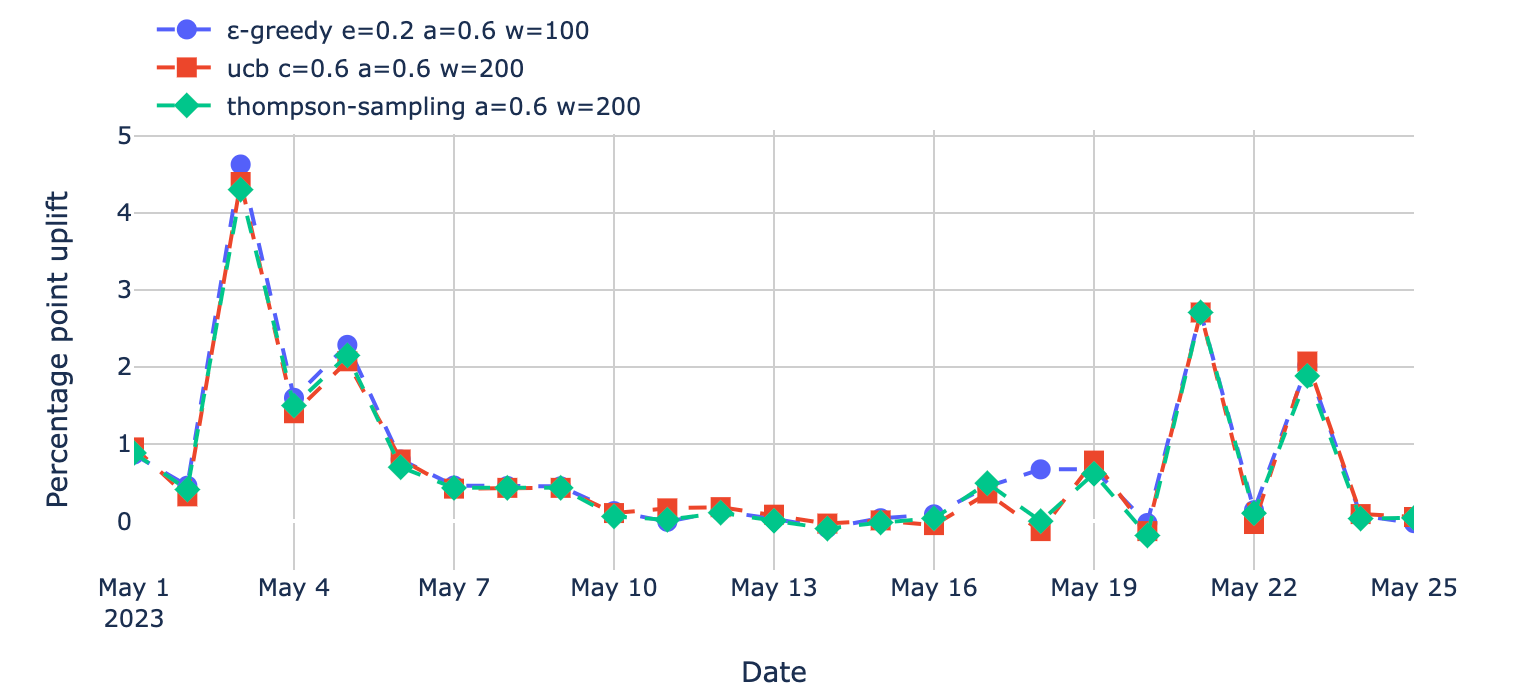}
  \caption{Online results: Uplift across Bandit Algorithms in production environment from May 1 2023 - May 25 2023}
  \label{fig:online}
\end{figure*}

\subsection{Offline simulation}
In the offline simulation, we estimated the success rate of each gateway at each time instant by taking the empirical success probability of the gateway using 25 transaction data before the time instant and 25 transactions after the time instant. On the "tuning dataset" we then simulated several non-stationary bandit algorithms such as sliding window and discounted versions of UCB algorithm \cite{auer2019achieving,wei2021nonstationary}, Thompson sampling-based algorithm \cite{trovo2020sliding}, Boltzmann-Gumbel exploration \cite{cesa2017boltzmann}, and $\epsilon$ greedy algorithms with different hyperparameters. 
The results in Figure \ref{fig:offline} show that the sliding window UCB (with a sliding window length of 200 transactions) and discounted UCB (with a discount factor of 0.6) algorithms have the lowest cumulative regrets. Discounted Boltzmann Gumbel algorithm also achieves comparable performance as discounted UCB algorithm.
During the testing, three algorithms -- UCB algorithm with window size 200, Thompson sampling with discount factor 0.6 and $\epsilon$ greedy algorithm with $\epsilon = 0.2$ --  yielded the lowest regret. We used these three algorithms for testing in the live experiments in April-May 2023.

\subsection{Online Experiments}  
To evaluate the performance of optimal algorithms, we ran online experiments on around 240M transactions over 60 days. The transactions were randomly assigned to one of the four routing algorithms -- the standard currently used rule-based algorithm and the three bandit-based algorithms chosen using the approach described in the previous section. The standard algorithm received approximately 10\% of the transactions, and each of the bandit algorithms processed approximately 30\% of the transactions daily.  Figure \ref{fig:online} depicts the differences in the daily success rates and the success rates obtained by each of the three bandit algorithms for the period of May 1, 2023 to May 25, 2023. Overall, sliding-window-UCB gave the best cumulative uplift of 0.92\% over the one month period which is statistically significant.

\section{Conclusion and Future Work} \label{sec:conc}
Given the ecosystem constraints and system requirements, we discuss the trade-offs in the three system design choices to deploy the bandit-based real-time payment Routing service. Further, we propose a Ray-based implementation of the Routing Service, a highly effective choice for payment processing in high-load applications. The architecture's balance of low latency, high throughput, scalability, and cost-effectiveness makes it well-suited for handling large transaction volumes. Using the system, we can infer the best arm using our non-stationary bandit approach within 5ms in the online implementation. Load testing demonstrated impressive performance, exceeding 10,000 transactions per second with P99 latency under 5ms.

This research paper also highlights the effectiveness of non-stationary bandit algorithms for continuous decision-making in optimizing transaction routing. The study applied these algorithms to Dream11's payment processing system, conducting offline simulations to identify the best-performing algorithms and their corresponding hyperparameters. Subsequently, two bandit algorithms and the $\epsilon$-greedy algorithm were deployed on the live payment system, outperforming the current rule-based approach under various time-varying environments by 0.92\%. We believe that the algorithmic technique developed and the system designed can be replicated across industries that serve a large number of user transactions in a short period of time.

During pre-processing of transactions, we observed the potential to predict payment gateway success rates with significant accuracy over the next 30 seconds, offering opportunities to enhance overall success rates in fluctuating situations further. Future work aims to address this by predicting time series change points to minimize regret and improve the system's performance even further as described in the \cite{mcdonald2023impatient} paper. Further, we did not consider the cost of transaction fee and the context of the user transacting on the platform. Optimization of routing based on these considerations will be conducted in our future work.

\newpage
\bibliographystyle{ACM-Reference-Format}
\bibliography{main,blogs}

\end{document}